\documentclass[runningheads]{llncs}
\usepackage[T1]{fontenc}

\usepackage{amsfonts}
\usepackage{amssymb}
\usepackage{amsmath}
\usepackage{booktabs}
\usepackage{multirow}
\usepackage{pifont}
\usepackage{cite}
\usepackage{graphicx,verbatim}
\begin{document}
\title{Deep Expert Injection for Anchoring Retinal VLMs with Domain-Specific Knowledge}
%
\author{Shuai Lu\inst{1,5} \and
Meng Wang\inst{2} \and
Jia Guo\inst{3} \and Jiawei Du\inst{1} \and Bo Liu\inst{4} \and Shengzhu Yang\inst{1} \and Weihang Zhang\inst{1} \and Huazhu Fu\inst{5}$^{\star}$ \and Huiqi Li\inst{1}$^{\star}$}
%
%
\institute{Beijing Institute of Technology, Beijing, China \and
National University of Singapore, Singapore \and  Tsinghua University, Beijing, China \and The Hong Kong Polytechnic University, Hong Kong \and Institute of High Performance Computing, Agency for Science,\\ Technology and Research, Singapore \\ $^{\star}$ Corresponding authors \\
}

\maketitle              
%
\newcommand{\modelname}{\textbf{EyExIn}}

\begin{abstract}

Large Vision Language Models (LVLMs) show immense potential for automated ophthalmic diagnosis. However, their clinical deployment is severely hindered by lacking domain-specific knowledge. In this work, we identify two structural deficiencies hindering reliable medical reasoning: 1) the Perception Gap, where general-purpose visual encoders fail to resolve fine-grained pathological cues (e.g., microaneurysms); and 2) the Reasoning Gap, where sparse visual evidence is progressively overridden by massive language priors in deeper transformer layers, leading to ungrounded hallucinations.
To bridge these gaps, we propose \textbf{EyExIn}, a data-efficient framework designed to anchor retinal VLMs with expert knowledge via a Deep Expert Injection mechanism.
Our architecture employs an Expert-Aware Dual-Stream encoding strategy that decouples visual representation into a general stream for anatomical context and a specialized expert stream for pathological semantics.
To ensure high-fidelity integration, we design a Semantic-Adaptive Gated Fusion module, which dynamically amplifies subtle lesion signals while filtering irrelevant background noise. 
Furthermore, we introduce Adaptive Deep Expert Injection to embed persistent "Vision Anchors" by integrating fused visual features as residual biases directly into intermediate LLM layers. 
This mechanism creates a visual shortcut that forces the reasoning stack to remain strictly grounded in visual evidence.
Extensive experiments across four benchmarks demonstrate that our model consistently outperforms massive proprietary systems. 
EyExIn significantly enhances domain-specific knowledge embedding and achieves state-of-the-art precision in ophthalmic visual question answering, advancing the development of trustworthy ophthalmic AI.

\keywords{Retinal Disease Diagnosis \and Vision-Language Models \and VQA}
\end{abstract}

\section{Introduction}

The emergence of Large Vision Language Models (LVLMs) has established a new paradigm in cross-modal interaction, offering transformative potential for automated healthcare \cite{wang2024qwen2,bai2025qwen3,liu2023visual,guo2025deepseek}. By effectively processing complex multimodal data and executing advanced cognitive reasoning, these systems are emerging as highly promising tools for clinical automation~\cite{chen2024towards,lin2025healthgpt,li2025eyecaregpt,llava_med,li2024integrated,liu2025gemex}. Due to their inherent reliance on the precise interpretation of intricate visual evidence, ophthalmology stands to benefit profoundly from this technological leap, particularly in tasks such as fundus report generation and visual question answering (VQA). However, despite their impressive fluency, current general-domain LVLMs are severely hampered by a critical flaw: \textit{lack of domain-specific knowledge embedding} \cite{hallucination_survey}. In clinical settings, this deficiency manifests in two critical ways: fabricating plausible but non-existent lesions (false positives), and more perilously, failing to identify subtle pathological cues, thereby misdiagnosing diseased patients as healthy. Such missed diagnoses can cause patients to lose crucial windows for early intervention, posing unacceptable safety risks and fundamentally undermining trust in AI-assisted diagnosis.

Existing alignment strategies typically rely on "brute-force" data scaling, such as massive instruction tuning \cite{liu2023visual,dai2023instructblip,zambrano2025clinically} or extensive RLHF \cite{yu2024rlhf,ouyang2022training}. This is highly impractical in ophthalmology, where expert-annotated fundus images are privacy-sensitive and prohibitively expensive to acquire. Confined to low-data regimes, standard parameter-efficient fine-tuning fails to optimally utilize limited visual evidence, exacerbating two structural deficiencies. 
First, \textbf{the Perception Gap}: General visual encoders pre-trained on natural images struggle to resolve fine-grained pathological features (e.g., tiny microaneurysms), passing ambiguous tokens to the LLM. 
Second, \textbf{the Reasoning Gap}: Sparse medical gradients fail to thoroughly rewire deep reasoning layers. Faced with visual-semantic uncertainty, the LLM falls back on massive pre-trained language priors (Prior Domination), bypassing weak visual evidence to fabricate clinical text.

To bridge these structural gaps under limited medical supervision, we propose \textbf{EyExIn}, a data-efficient framework designed to deeply embed ophthalmic knowledge into LVLMs. Our main contributions are summarized as follows:
\begin{itemize}
    \item To tackle the \textit{Perception Gap}, we introduce an \textbf{Expert-Aware Dual-Stream architecture} with a \textbf{Semantic-Adaptive Gated Fusion} module. It dynamically decouples global anatomical context from fine-grained expert pathology, filtering background noise to maximize the visual Signal-to-Noise Ratio of subtle lesions.
    \item  To overcome the \textit{Reasoning Gap}, we propose an \textbf{Adaptive Deep Expert Injection} mechanism. By embedding visual features as persistent residual biases directly into intermediate LLM layers, our framework establishes "Vision Anchors" that strictly ground pathological predictions in actual image evidence.
    \item Extensive experiments across four benchmarks (TM4K, JSIEC, Retina, and ODIR) demonstrate that our  \textbf{EyExIn} with 7B-parameter achieves state-of-the-art diagnostic precision. 
\end{itemize}

\section{Methodology}

\begin{figure*}[!t]
    \centering
    \includegraphics[width=1.0\linewidth]{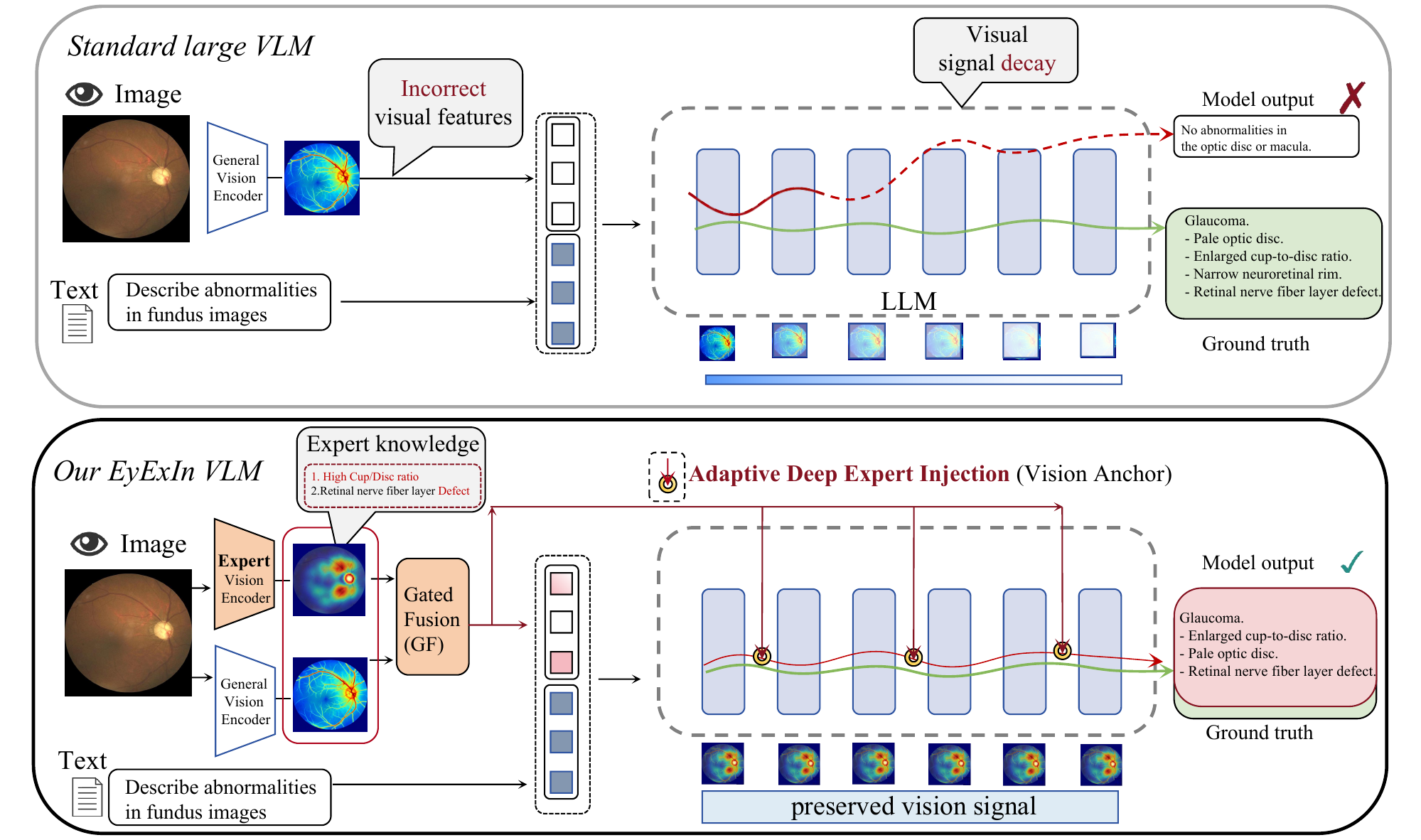}
    \caption{Architectural comparison between a standard VLM and the proposed EyExIn framework. \textbf{(Top)} Standard VLMs suffer from visual signal decay across deep LLM layers, leading to ungrounded reasoning. \textbf{(Bottom)} EyExIn introduces an expert-aware dual-stream encoder and Semantic-Adaptive Gated Fusion (GF). Crucially, Deep Expert Injection establishes persistent ``Vision Anchors'' in intermediate layers, ensuring evidence-grounded clinical outputs.}
    \label{fig:architecture}
\end{figure*}

The overall architecture of \modelname\ is illustrated in Fig.~\ref{fig:architecture}. Compared to standard early-fusion VLMs (Top in Fig.~\ref{fig:architecture}) characterized by coarse perception and visual signal decay across deep LLM layers, our framework introduces a decoupled, deep-injected paradigm (Bottom in Fig.~\ref{fig:architecture}). The data flow is explicitly structured to target two structural deficiencies. First, to address the \textit{Perception Gap}, the input fundus image is processed by an Expert-Aware Dual-Stream encoder, followed by a Semantic-Adaptive Gated Fusion (GF) module designed to isolate fine-grained lesions from the anatomical background. Second, to address the \textit{Reasoning Gap}, \modelname\ bypasses traditional prompt-level visual integration. Instead, an Adaptive Deep Expert Injection mechanism embeds the fused features directly into intermediate LLM layers as persistent "Vision Anchors". 

\subsection{Expert-Aware Dual-Stream Encoding}
Standard LVLMs \cite{yu2025unifying,liu2023visual,Bai2025qwen2_5} relying on single natural-image encoders struggle to resolve fine-grained pathologies in medical domains. To bridge this \textit{Perception Gap}, we decouple visual extraction into two complementary streams. 

\textbf{General Stream (Anatomical Context):} We retain the frozen foundation encoder (e.g., Qwen2.5-VL\cite{Bai2025qwen2_5}) to extract global features $F_{gen} \in \mathbb{R}^{N \times D_{llm}}$ ($N$ tokens). Leveraging its broad pre-training, it preserves macroscopic anatomical structures and holistic colorimetric variations (e.g., optic disc pallor) that are often compromised in heavily specialized models.

\textbf{Expert Stream (Pathological Semantics):} A contrastively pre-trained fundus foundation encoder \cite{guo2025native} extracts fine-grained features $F_{exp} \in \mathbb{R}^{N \times D_{exp}}$, providing high sensitivity to subtle lesions like microaneurysms. To align the feature spaces, a linear projection $\phi$ maps the expert features to the general dimension, yielding $F'_{exp} = \phi(F_{exp}) \in \mathbb{R}^{N \times D_{llm}}$.

\subsection{Semantic-Adaptive Gated Fusion}
Trivial integration strategies (e.g., direct addition or concatenation) are sub-optimal as they dilute fragile lesion signals, inject expert-driven noise into healthy regions, or impose severe computational redundancy. Furthermore, relying solely on the expert encoder discards the general stream's critical sensitivity to holistic colorimetric attributes. To isolate subtle lesions while preserving global anatomy, we propose a Semantic-Adaptive Gated Fusion, which employs a lightweight semantic router to compute a token-wise weight map $\alpha \in \mathbb{R}^{N \times 1}$:
\begin{equation}
\alpha = \sigma(\text{MLP}([F_{gen} \parallel F'_{exp}])),
\end{equation}
where $\sigma(\cdot)$ denotes the Sigmoid activation function and $\parallel$ represents channel concatenation. The fused representation is obtained via convex interpolation:
\begin{equation}
F_{fused} = (1 - \alpha) \odot F_{gen} + \alpha \odot F'_{exp}.
\end{equation}
By dynamically enforcing $\alpha \to 1$ in pathological regions and $\alpha \to 0$ in broader structural contexts, this spatial adaptivity purifies the visual features and maximizes the Signal-to-Noise Ratio.

\subsection{Adaptive Deep Expert Injection}
In standard early fusion, fragile visual signals decay across deep reasoning layers, causing the model to fabricate diagnoses based on language priors. While recent cross-layer fusions (e.g., Qwen3-VL \cite{bai2025qwen3}) mitigate this via uniform residual additions, clinical applications can be further optimized by a more targeted paradigm that distinguishes critical lesions from normal anatomical backgrounds. To establish a strict and focused visual grounding, we directly inject the expert-fused features $F_{fused} \in \mathbb{R}^{N \times D_{llm}}$ back into the corresponding visual token positions within the intermediate LLM layers.

Crucially, this integration must safely supplement visual knowledge without erasing the LLM's pre-trained hidden states. For the visual tokens at the $l$-th Transformer layer, denoted as $H^{vis}_{l-1} \in \mathbb{R}^{N \times D_{llm}}$, we propose an \textbf{Adaptive Deep Expert Injection}. Rather than blind addition, we first compute a token-wise spatial routing map $g_l = \sigma(\text{MLP}_l([H^{vis}_{l-1} \parallel F_{fused}])) \in \mathbb{R}^{N \times 1}$. By concatenating the current hidden states with the pristine visual evidence, this router explicitly detects representation decay, adaptively evaluating the necessity of visual refreshing. The expert features are then integrated as a persistent residual bias:

\begin{equation}
    H'^{vis}_l = H^{vis}_{l-1} + \tanh(\gamma_l) \cdot (g_l \odot F_{fused})
\end{equation}
where $\gamma_l$ is a zero-initialized, layer-specific scaling parameter. This dual-gated design explicitly decouples the feature integration process. At the token level, $g_l$ functions as a discrepancy-aware router that selectively bypasses normal anatomical backgrounds, thereby precisely refreshing pathological representations when language priors risk overriding visual evidence. At the layer level, $\tanh(\gamma_l)$ acts as a global capacity controller; its zero-initialization effectively isolates the pre-trained LLM from uncalibrated visual projections during early training, ensuring robust convergence and averting catastrophic forgetting.

\section{Experiments and Results}

\paragraph{\textbf{Experimental Datasets.}} To ensure a fair comparison, all fine-tuned models are trained on a real-world clinical dataset of 150K fundus images and their corresponding diagnostic reports. We evaluate its diagnostic generalization across four diverse datasets covering various retinal diseases and imaging devices. This includes a large-scale private clinical dataset (\textbf{TM4K}: 4,022 images spanning 21 categories) and three public datasets: \textbf{JSIEC} \cite{cen2021automatic} (315 images with 39 classes), \textbf{Retina} \cite{Retina} (90 images with 3 classes), and \textbf{ODIR} \cite{odir} (1,851 images with 7 classes).

\paragraph{\textbf{Implementation Details.}}We employ Qwen2.5-VL (7B) \cite{Bai2025qwen2_5} as our baseline. We freeze its general vision encoder while introducing a pre-trained fundus foundation model \cite{guo2025native} as the trainable expert stream. To achieve parameter-efficient optimization under limited data, the LLM backbone is fine-tuned using Low-Rank Adaptation (LoRA) with a rank of 16 and a scaling factor ($\alpha$) of 32. Training is conducted on A100 GPUs via DeepSpeed ZeRO-2, utilizing the AdamW optimizer with a weight decay of 0.05, a peak learning rate of $2 \times 10^{-5}$, and a cosine learning rate decay schedule incorporating a 3\% warmup. The global batch size is set to 48, and the maximum sequence length is constrained to 2048 tokens. Furthermore, employing BFloat16 mixed-precision format guarantees both memory efficiency and computational stability during the training process.

We compare our 7B-model against proprietary LVLMs (\textbf{Qwen3-VL-Max}, \textbf{ChatGPT-5.2}, and \textbf{Gemini3-Pro}) and open-source models (\textbf{LLaVA}, and \textbf{Qwen2.5-VL}) fine-tuned (FT) on our data. Diagnostic performance is evaluated via Macro-Averaged Recall, Precision, and F1-score, rigorously ensuring rare pathologies contribute equally. Additionally, to assess the linguistic and clinical fidelity of generated text, we utilize BLEU-1, ROUGE-L, and METEOR for structural similarity, alongside BERT-F1 for deep semantic equivalence.

\begin{table*}[!t]
\caption{Comparison of \textbf{Macro Average} metrics on \textbf{multi-class} Closed and Open-ended VQA tasks. Best results are highlighted in \textbf{bold}. Models are categorized as \textbf{Proprietary (Prop.)} general-domain LVLMs and \textbf{Fine-Tuned (FT)} open-source models.}
\label{tab:merged_qa_vqa_wide}
\centering
  \scriptsize
  
  \setlength{\tabcolsep}{6pt}
  
  \begin{tabular}{@{}ll ccc | ccc@{}}
    \toprule
    \multirow{2}{*}{\textbf{Dataset}} & \multirow{2}{*}{\textbf{Method}} & \multicolumn{3}{c}{\textbf{Closed VQA (\%)}} & \multicolumn{3}{c}{\textbf{Open-ended VQA (\%)}} \\
    \cmidrule(lr){3-5} \cmidrule(lr){6-8}
    & & F1 & \shortstack{Recall} & Prec & F1 & \shortstack{Recall} & Prec \\
    \midrule

    \multirow{6}{*}{\textbf{TM4K}} 
    & Qwen3-VL-Max (Prop.) & 7.82 & 9.16 & 14.58 & 11.39 & 11.97 & 32.58 \\
    & ChatGPT-5.2 (Prop.)  & 15.93 & 21.55 & 23.83 & 26.72 & 27.44 & 44.20 \\
    & Gemini3-Pro (Prop.)  & 37.99 & 48.11 & 40.53 & 47.71 & 54.17 & 53.25 \\
    & LLaVA \cite{liu2023visual} (FT)            & 39.83 & 46.13 & 44.80 & 39.71 & 43.01 & 44.61 \\
    & Qwen2.5-VL \cite{Bai2025qwen2_5} (FT)      & 55.72 & 69.12 & 51.22 &  47.84 & 56.02 & 54.44 \\
    & \textbf{Our \modelname} (FT) & \textbf{78.07} & \textbf{82.42} & \textbf{77.33} & \textbf{72.91} & \textbf{78.99} & \textbf{71.87} \\
    \midrule

    \multirow{6}{*}{\textbf{JSIEC}} 
    & Qwen3-VL-Max (Prop.) & 6.96 & 9.55 & 7.59 &  19.48 & 24.34 & 24.98 \\
    & ChatGPT-5.2 (Prop.)  & 21.18 & 26.00 & 26.09 & 31.39 & 35.80 & 38.86 \\
    & Gemini3-Pro (Prop.)  & 43.29 & 49.27 & 48.25 & 30.78 & 26.34 & 48.16 \\
    & LLaVA \cite{liu2023visual}(FT)            &  26.59 & 33.62 & 30.52 & 11.82 & 12.15 & 17.65 \\
    & Qwen2.5-VL \cite{Bai2025qwen2_5}(FT)      & 44.98 & 47.52 &  53.29 & 14.00 & 14.85 & 21.99 \\
    & \textbf{Our \modelname} (FT) & \textbf{80.66} & \textbf{82.33} & \textbf{85.20} & \textbf{63.10} & \textbf{76.32} & \textbf{60.84} \\
    \midrule

    \multirow{6}{*}{\textbf{Retina}} 
    & Qwen3-VL-Max (Prop.) & 42.00 & 41.36 & 49.71 & 53.00 & 48.77 & 73.74 \\
    & ChatGPT-5.2 (Prop.)  & 56.63 & 61.73 & 62.15 & 60.28 & 51.23 & 87.66 \\
    & Gemini3-Pro (Prop.)  & 65.73 & \textbf{71.60} & 64.63 & 61.42 & 45.68 & 95.83 \\
    & LLaVA \cite{liu2023visual}(FT)            & 47.60 & 55.42 & 43.11 & 45.89 & 54.30 & 41.55 \\
    & Qwen2.5-VL\cite{Bai2025qwen2_5} (FT)      & 54.59 & 63.58 & 48.15 & 52.63 & 61.73 & 46.90 \\
    & \textbf{Our \modelname} (FT) & \textbf{71.27} & 67.90 & \textbf{89.68} & \textbf{67.80} & \textbf{62.30} & \textbf{96.15} \\
    \midrule

    \multirow{6}{*}{\textbf{ODIR}} 
    & Qwen3-VL-Max (Prop.) & 29.65 & 27.07 & 43.65 & 31.65 & 30.73 & 53.58 \\
    & ChatGPT-5.2 (Prop.)  & 29.06 & 31.08 & 33.19 & 36.42 & 36.73 & 42.54 \\
    & Gemini3-Pro (Prop.)  & 29.48 & 40.39 & 34.56 & 32.13 & 24.54 & \textbf{75.70} \\
    & LLaVA \cite{liu2023visual}(FT)            & 37.82 & 39.55 & 38.41 & 34.90 & 36.15 & 40.80 \\
    & Qwen2.5-VL \cite{Bai2025qwen2_5}(FT)      & 42.84 & 44.87 & 44.69 & 39.09 & 42.02 & 46.28 \\
    & \textbf{Our \modelname} (FT) & \textbf{60.09} & \textbf{59.81} & \textbf{64.69} & \textbf{56.70} & \textbf{55.20} & 60.40 \\
    \bottomrule
  \end{tabular}
\end{table*}

\paragraph{\textbf{Main Results: Visual Question Answering.}}
\label{sec:vqa_results}
Table~\ref{tab:merged_qa_vqa_wide} and Table~\ref{tab:tm4k_text_similarity} present the comprehensive evaluation. As observed, general proprietary LVLMs struggle to provide accurate diagnostic assessments, exhibiting remarkably low F1-scores in Closed VQA (e.g., 7.82\% for Qwen3-VL-Max and 15.93\% for ChatGPT-5.2 on TM4K) due to their inability to resolve fine-grained ophthalmic features. While standard domain-specific fine-tuning (LLaVA, Qwen2.5-VL) partially mitigates this perception gap, \textbf{\modelname} demonstrates consistently superior capabilities. Focusing on Closed VQA, our framework achieves state-of-the-art F1-scores of 78.07\% on TM4K and 80.66\% on JSIEC. This success is directly attributed to our Semantic-Adaptive Gated Fusion, which effectively isolates pathological cues to drive high diagnostic sensitivity (e.g., 82.42\% Recall on TM4K). Furthermore, by employing Adaptive Deep Expert Injection to anchor reasoning strictly to visual evidence, \modelname\ effectively suppresses ungrounded language priors. This advantage seamlessly extends to the highly challenging Open-ended VQA task, where the model maintains remarkable precision (e.g., 96.15\% on Retina) compared to the severe false-positive drops of baseline models. This strict clinical fidelity is further corroborated by Table~\ref{tab:tm4k_text_similarity}, where \modelname\ leads across all structural and semantic text generation metrics, objectively validating our targeted architectural design in low-data regimes.

\begin{table*}[!t]
  \caption{Comparison of text similarity metrics on the \textbf{TM4K dataset}. }
  \label{tab:tm4k_text_similarity}
  \centering
  \scriptsize
  
  \setlength{\tabcolsep}{6pt}
  \begin{tabular}{l cccc}
    \toprule
     \textbf{Method} & \textbf{BLEU-1} & \textbf{ROUGE-L} & \textbf{METEOR} & \textbf{BERT-F1} \\
    \midrule
    
     Qwen3-VL-Max & 0.3623 & 0.2082 & 0.3258 & 0.9352 \\
     ChatGPT-5.2 & 0.2942 & 0.2098 & 0.2726 & 0.9320 \\
     Gemini3-Pro & 0.2564 & 0.2130 & 0.3056 & 0.9286 \\
     LLaVA \cite{liu2023visual}(FT) & 0.3854 & 0.2688 & 0.3692 & 0.9387 \\
     Qwen2.5-VL \cite{Bai2025qwen2_5}(FT) & 0.4123 & 0.3046 & 0.4076 & 0.9420 \\
     \textbf{Our EyExIn} (FT) & \textbf{0.4579} & \textbf{0.3552} & \textbf{0.4316} & \textbf{0.9466} \\
    \bottomrule
  \end{tabular}
\end{table*}

\begin{figure}[!t]
    \centering
    \includegraphics[width=1\linewidth]{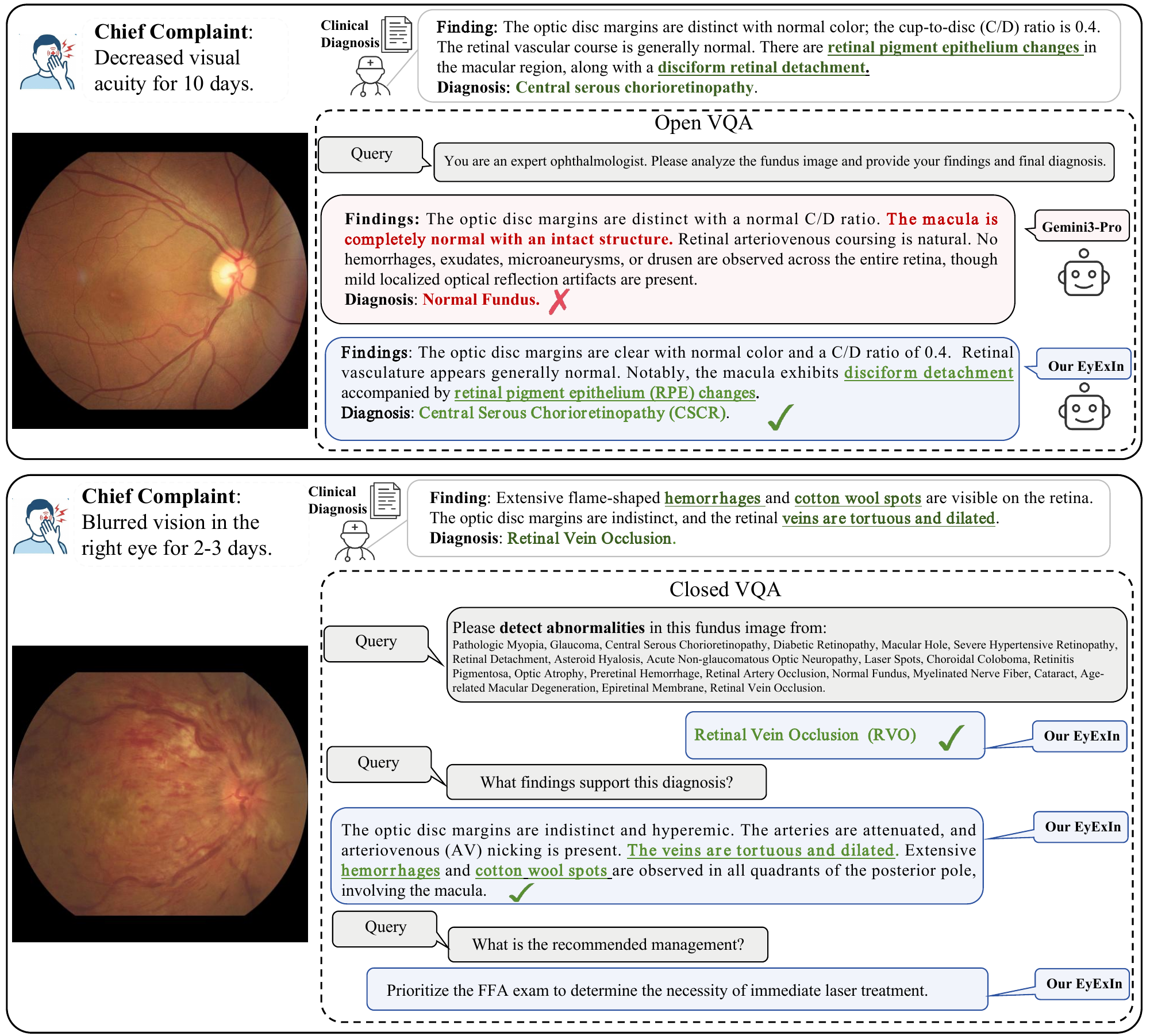}
\caption{Qualitative VQA comparison on two real-world clinical cases. Red text indicates severe missed diagnoses and hallucinated artifacts by Gemini, while green text highlights accurate, expert-aligned lesion identification and clinical metrics by EyExIn.}
    \label{fig:qualitative} 
\end{figure}

\textbf{Real-World Clinical Validation:} Fig.~\ref{fig:qualitative} demonstrates \modelname's reliability across Central Serous Chorioretinopathy (CSCR) and Retinal Vein Occlusion (RVO) cases. In Open VQA, Gemini3-Pro miss subtle detachments and hallucinate a Normal Fundus diagnosis. Conversely, driven by persistent visual anchors, \modelname\ accurately extracts quantitative metrics (e.g., C/D ratio 0.4) for CSCR. Furthermore, in Closed VQA, it correctly identifies RVO by strictly grounding its reasoning in multi-regional lesions (tortuous veins, extensive hemorrhages), demonstrating its capacity for trustworthy, evidence-based clinical evaluation.

\begin{table}[!t]
\caption{Ablation study on the TM4K dataset (Closed VQA). We compare different strategies for integrating the expert stream: sequence addition (Add) vs. our semantic-adaptive gated fusion (Gated), and unconditional addition (Direct) vs. our adaptive routing (Adapt) for deep injection.}
  \label{tab:ablation}
\centering
  \scriptsize
  \setlength{\tabcolsep}{5.5pt}  
  \begin{tabular}{@{}l ccc | ccc@{}}
    \toprule
    \textbf{Model} & \textbf{Expert} & \textbf{Fusion} & \textbf{Injection} & \multirow{2}{*}{\textbf{F1}} & \multirow{2}{*}{\textbf{Recall}} & \multirow{2}{*}{\textbf{Prec}} \\
    \textbf{Variant} & \textbf{Stream} & \textbf{Strategy} & \textbf{Strategy} & & & \\
    \midrule
    Baseline (Qwen2.5-VL)      & - & - & - & 55.72 & 69.12 & 51.22 \\
    Variant 1           & \checkmark & Add & - & 64.23 & 81.30 & 55.10 \\
    Variant 2           & \checkmark & Gated & - & 71.45 & 78.50 & 67.65 \\
    Variant 3           & \checkmark & Gated & Direct & 74.12 & 82.15 & 69.80 \\
    \textbf{\modelname} & \checkmark & \textbf{Gated} & \textbf{Adapt} & \textbf{78.07} & \textbf{82.42} & \textbf{77.33} \\
    \bottomrule
  \end{tabular}
\end{table}

\paragraph{\textbf{Ablation Study.}}
To validate the proposed components, we conduct an ablation study on the closed validation set of TM4K as summarized in Table~\ref{tab:ablation}.

\textbf{Expert Stream \& Gated Fusion:} Integrating the expert stream via simple element-wise addition (Variant 1) significantly boosts Recall (from 69.12\% to 81.30\%), confirming its ability to capture fine-grained pathological cues missed by the general encoder. However, this naive integration injects uncalibrated noise into healthy regions, bottlenecking Precision at 55.10\%. Upgrading to Semantic-Adaptive Gated Fusion (Variant 2) dynamically filters this noise, yielding a substantial Precision leap (+12.55\%) by maximizing the visual signal-to-noise ratio before LLM processing.

\textbf{Adaptive Deep Expert Injection:} To prevent visual signal decay in deeper reasoning layers, Variant 3 unconditionally adds visual features (Direct). While this forces visual grounding and further boosts Recall (82.15\%), it indiscriminately alters grammatical tokens, disrupting the LLM's pre-trained syntactic fluency and capping Precision at 69.80\%. Our \modelname\ resolves this structural conflict via adaptive routing (Adapt). By intelligently bypassing visual signals for grammatical tokens while strictly anchoring pathological predictions, it drives a massive gain in Precision (77.33\%) and achieves the optimal F1-score (78.07\%).

\section{Conclusion}
In this paper, we presented \textbf{EyExIn}, a novel framework designed to embed domain-specific knowledge into retinal Large Vision Language Models (LVLMs). By decoupling visual perception into specialized anatomical and pathological streams, our architecture effectively resolves the Perception Gap that limits general purpose encoders.
Our framework utilizes a dual-stream architecture with Semantic-Adaptive Gated Fusion to accurately isolate subtle lesions while preserving global anatomical context. Furthermore, Deep Expert Injection embeds persistent visual anchors into the LLM's deep layers, effectively preventing visual signal decay. 
Extensive experiments across four diverse fundus benchmarks demonstrate that our 7B parameter model achieves state-of-the-art diagnostic precision in both Closed and Open-ended VQA tasks, consistently outperforming massive proprietary systems such as GPT-5.2 and Gemini3-Pro. By strictly grounding clinical reasoning in high fidelity image evidence, EyExIn advances the development of trustworthy and evidence based AI for ophthalmology.

\bibliographystyle{splncs04}

\bibliography{ref}

\end{document}